\documentclass[sigconf]{acmart}

\AtBeginDocument{%
  \providecommand\BibTeX{{%
    \normalfont B\kern-0.5em{\scshape i\kern-0.25em b}\kern-0.8em\TeX}}}

\copyrightyear{2023}
\acmYear{2023}
\setcopyright{acmlicensed}\acmConference[GECCO '23]{Genetic and Evolutionary Computation Conference}{July 15--19, 2023}{Lisbon, Portugal}
\acmBooktitle{Genetic and Evolutionary Computation Conference (GECCO '23), July 15--19, 2023, Lisbon, Portugal}
\acmPrice{15.00}
\acmDOI{10.1145/3583131.3590358}
\acmISBN{979-8-4007-0119-1/23/07}

\usepackage{latexsym}
\usepackage{amsmath} 
\usepackage{mathtools}       
\usepackage{stmaryrd}        
\usepackage{bm}              
\DeclareMathOperator{\Tr}{Tr}     
\DeclareMathOperator{\Cov}{Cov}   
\DeclareMathOperator{\vect}{vec}  

\providecommand{\E}{\mathbb{E}} 
\providecommand{\T}{\mathrm{T}} 
\renewcommand{\leq}{\leqslant} 
\DeclarePairedDelimiterX{\inner}[2]{\langle}{\rangle}{#1, #2}
\DeclarePairedDelimiter{\norm}{\lVert}{\rVert}

\usepackage{amsthm}
\usepackage{xcolor}
\usepackage{ifthen}
\usepackage{cleveref}

\theoremstyle{definition}

\usepackage{algorithm}
\usepackage{algpseudocode}

\algnewcommand\algorithmicset{\textbf{Set:}}
\algnewcommand\Set{\item[\algorithmicset]}

\providecommand{\mvae}{\mathcal{E}}
\providecommand{\mvav}{\mathcal{V}}
\providecommand{\etam}{\eta_{m}}
\providecommand{\etasig}{\eta_{\Sigma}}

\makeatletter
\let\OldStatex\Statex
\renewcommand{\Statex}[1][3]{%
  \setlength\@tempdima{\algorithmicindent}%
  \OldStatex\hskip\dimexpr#1\@tempdima\relax}
\makeatother

\begin{document}

\title{CMA-ES with Learning Rate Adaptation: Can CMA-ES with Default Population Size Solve Multimodal and Noisy Problems?}

\author{Masahiro Nomura}
\authornote{Corresponding author}
\email{nomura.m.ad@m.titech.ac.jp}
\orcid{0000-0002-4945-5984}
\affiliation{%
  \institution{Tokyo Institute of Technology}
  \streetaddress{4259 Nagatsutach\={o}, Midori Ward}
  \city{Yokohama}
  \state{Kanagawa}
  \country{Japan}
  \postcode{226-0026}
}
\author{Youhei Akimoto}
\email{akimoto@cs.tsukuba.ac.jp}
\orcid{0000-0003-2760-8123}
\affiliation{%
  \institution{University of Tsukuba \& RIKEN AIP}
  \streetaddress{1-1-1 Tennodai}
  \city{Tsukuba}
  \state{Ibaraki}
  \country{Japan}
  \postcode{305-8573}
}

\author{Isao Ono}
\email{isao@c.titech.ac.jp}
\orcid{0009-0008-2110-9853}
\affiliation{%
  \institution{Tokyo Institute of Technology}
  \streetaddress{4259 Nagatsutach\={o}, Midori Ward}
  \city{Yokohama}
  \state{Kanagawa}
  \country{Japan}
  \postcode{226-0026}
}


\begin{abstract} 
The covariance matrix adaptation evolution strategy (CMA-ES) is one of the most successful methods for solving black-box continuous optimization problems.
One practically useful aspect of the CMA-ES is that it can be used without hyperparameter tuning.
However, the hyperparameter settings still have a considerable impact, especially for \emph{difficult} tasks such as solving multimodal or noisy problems.
In this study, we investigate whether the CMA-ES with default population size can solve multimodal and noisy problems.
To perform this investigation, we develop a novel learning rate adaptation mechanism for the CMA-ES, such that the learning rate is adapted so as to maintain a constant signal-to-noise ratio.
We investigate the behavior of the CMA-ES with the proposed learning rate adaptation mechanism through numerical experiments, and compare the results with those obtained for the CMA-ES with a fixed learning rate.
The results demonstrate that, when the proposed learning rate adaptation is used, the CMA-ES with default population size works well on multimodal and/or noisy problems, \emph{without} the need for extremely expensive learning rate tuning.
\end{abstract}



\maketitle

\section{Introduction}
\label{sec:intro}
Among the existing methods for solving black-box continuous optimization problems, the covariance matrix adaptation evolution strategy (CMA-ES)~\cite{hansen2001completely,hansen2016cma} is one of the most successful. In the CMA-ES, optimization is performed by updating the multivariate normal distribution. That is, the CMA-ES first samples candidate solutions from the distribution and then updates the distribution parameters (i.e., the mean vector $m$ and covariance matrix $\Sigma = \sigma^2 C$) based on the objective function $f$ value.
The CMA-ES update is partly based on the natural gradient descent~\cite{akimoto2010bidirectional,ollivier2017information} of the expected $f$, and
$m$ and $C$ in the CMA-ES are updated to decrease the expected evaluation value.
One practically useful aspect of the CMA-ES is that it is a quasi-hyperparameter-free algorithm; that is,
practitioners can use the CMA-ES without tuning its hyperparameters, because default values are provided for all hyperparameters based on theoretical analysis and extensive empirical evaluations. In detail, the hyperparameter values are automatically computed from the dimension $d$ and population size $\lambda$ where, by default, $\lambda = 4 + \lfloor 3 \ln(d) \rfloor$.

While the default $\lambda$ works well for a wide range of unimodal problems, increasing $\lambda$ from this default value can be helpful for \emph{difficult} tasks such as solving multimodal and additive noisy problems~\cite{hansen2004evaluating,psaigo,psacmaes}.
However, in a black-box scenario, determining the problem structure of $f$ is challenging and, thus, determining the appropriate $\lambda$ in advance is similarly challenging.
To resolve this issue, online adaptation of $\lambda$ has been proposed~\cite{psaigo,psacmaes,pccmsaes,cmaesapop,loshchilov2014maximum}.
The population size adaptation (PSA)-CMA-ES~\cite{psacmaes} is a representative $\lambda$ adaptation mechanism with promising performance when applied to difficult tasks, including multimodal and additive noisy problems.

It has been observed that, in the CMA-ES, increasing $\lambda$ has an effect similar to decreasing the $m$ learning rate, i.e., $\eta_m$~\cite{cmasmallcm}\footnote{Note that, in Ref.~\cite{cmasmallcm}, the rank-one update was excluded from the CMA-ES. In this study, however, we consider the CMA-ES that includes the rank-one update. As an additional note, in \cite{cmasmallcm}, the notation $c_m$ is used instead of $\eta_m$.}. Indeed, the $m$ and $\Sigma$ learning rate, i.e., $\eta$, is another hyperparameter that critically impacts performance.
If $\eta$ is too high, the parameter update is unstable; however, an excessively small $\eta$ degrades the search efficiency.
Miyazawa and Akimoto~\cite{cmasmallcm} reported that the CMA-ES with even a relatively small $\lambda$ (e.g., $\lambda = \sqrt{d}$) solves multimodal problems, through appropriate setting of $\eta$.
However, discovery of such an appropriate $\eta$ is difficult in practice, as prior knowledge is often limited and hyperparameter tuning requires extremely expensive numerical investigations.
Therefore, the realization of online $\eta$ adaptation depending on the problem difficulty will constitute an important advance, as practitioners can then use the CMA-ES \emph{safely}, without the need for prior knowledge or expensive trial-and-error calculations. In particular,
we believe that $\eta$ adaptation has advantages over $\lambda$ adaptation from a practical perspective, as the former is more suited to parallel implementation.
For example, practitioners often wish to specify a certain number of workers as the value of $\lambda$, to avoid wasting computing resources.
However, $\lambda$ adaptation may not fully utilize the available computing resources, depending on the variations in this value that occur during optimization.
In contrast, with $\eta$ adaptation, the available resources can be fully exploited, as $\lambda$ is fixed to the maximum number of workers.
Moreover, with $\eta$ adaptation, the parameter update is regularly performed, whereas the CMA-ES with $\lambda$ adaptation produces no progress until all $\lambda$ solutions are evaluated. 
This fact makes it difficult to determine when to terminate the search.

Online adaptation of $\eta$ itself is not new, and several previous studies attempted to adapt the $\eta$ values in CMA-ES variants; however, those adaptations targeted \emph{speed-up}~\cite{lradaptnes,gissler2022learning,loshchilov2014maximum}.
One notable exception is the $\eta$ adaptation proposed by Krause~\cite{noiseresilientes}, which aims to solve additive noisy problems by introducing new evolution strategies\footnote{Note that there are already ESs that use a (fixed) $\eta$ smaller than one (i.e., rescaled mutations)~\cite{Rec94,Beyer98,Beyer00}.}.
However, in this approach, the problem difficulty is estimated through resampling, i.e., by repeating an evaluation of the \emph{same} solution; thus, it is not suitable for solving (noiseless) multimodal problems.
Furthermore, the internal parameters of the evolution strategies are greatly modified and, thus, their direct application to the CMA-ES is difficult.

The aim of this study is to determine whether the CMA-ES with default $\lambda$ can solve multimodal and additive noisy problems without the need for extremely expensive $\eta$ tuning, and without adjusting the CMA-ES apart from $\eta$.
To achieve this, we propose an $\eta$ adaptation mechanism for the CMA-ES, which we call the learning rate adaptation (LRA)-CMA-ES, through which $\eta$ is adapted to maintain a constant signal-to-noise ratio (SNR).
The key feature of the proposed method is that specific knowledge of the internal mechanism of the distribution parameter update is not required in order to estimate the SNR.
As a result, the proposed method is widely applicable to different CMA-ES variants (e.g., diagonal decoding (dd)-CMA~\cite{ddcma}). This is despite the fact that the present study considers the most commonly used CMA-ES, which combines the weighted recombination, step-size $\sigma$ adaptation, rank-one update, and rank-$\mu$ update.

The remainder of this paper is organized as follows:
Section~\ref{sec:cma} explains the CMA-ES algorithm, Section~\ref{sec:lra} presents the proposed $\eta$ adaptation mechanism based on the SNR estimation, Section~\ref{sec:exp} evaluates the proposed $\eta$ adaptation on noiseless and noisy problems and, finally, Section~\ref{sec:conclusion} concludes with a summary and a discussion of future work.

\section{CMA-ES}
\label{sec:cma}

We consider minimization of the objective function $f: \mathbb{R}^d \to \mathbb{R}$.
The CMA-ES uses a multivariate normal distribution to generate candidate solutions, where the $\mathcal{N}(m, \sigma^2 C)$ distribution is parameterized by three elements: the mean vector $m \in \mathbb{R}^d$, the step-size $\sigma \in \mathbb{R}_{>0}$, and the covariance matrix $C \in \mathbb{R}^{d\times d}$.

The CMA-ES first initializes the $m^{(0)}, \sigma^{(0)}$, and $C^{(0)}$ parameters.
Then, the following steps are repeated until a pre-defined stopping criterion is satisfied.

\noindent
{\bf Step 1. Sampling and Evaluation}\\
At iteration $t + 1$ (where $t$ begins at $0$), $\lambda$ candidate solutions $x_i\ (i=1, 2, \cdots, \lambda)$ are sampled independently from $\mathcal{N}(m^{(t)}, ( \sigma^{(t)} )^2 C^{(t)})$, as follows:
\begin{align}
    y_i &= \sqrt{ C^{(t)} } z_i, \\
    x_i &= m^{(t)} + \sigma^{(t)} y_i,
\end{align}
where $z_i \sim \mathcal{N}(0, I)$, with $I$ being the identity matrix.
The solutions are evaluated on $f$ and sorted in ascending order.
We let $x_{i:\lambda}$ be the $i$-th best candidate solution $f(x_{1:\lambda}) \leq f(x_{2:\lambda}) \leq \cdots \leq f(x_{\lambda:\lambda})$.
Additionally, we let $y_{i:\lambda}$ and $z_{i:\lambda}$ be random vectors corresponding to $x_{i:\lambda}$.

\noindent
{\bf Step 2. Computation of Evolution Paths}\\
Using the weight function $w_i$ and the parent number $\mu \leq \lambda$, the weighted averages $dy = \sum_{i=1}^{\mu} w_i y_{i:\lambda}$ and $dz = \sum_{i=1}^{\mu} w_i z_{i:\lambda}$ are calculated.
Then, the evolution paths are updated as follows:
\begin{align}
    p_{\sigma}^{(t+1)} &= (1-c_{\sigma}) p_{\sigma}^{(t)} + \sqrt{c_{\sigma} (2-c_{\sigma}) \mu_w} dz, \\
    p_{c}^{(t+1)} &= (1-c_{c}) p_{c}^{(t)} + h_{\sigma}^{(t+1)} \sqrt{c_{c} (2-c_{c}) \mu_w} dy,
\end{align}
where $\mu_w = 1 / \sum_{i=1}^{\mu} w_i^2$, $c_{\sigma}$, and $c_c$ are the cumulation factors, and $h_{\sigma}^{(t+1)}$ is the Heaviside function, which is defined as follows~\cite{hansen2014principled}:
\begin{align}
    h_{\sigma}^{(t+1)} = \begin{cases}
1 & {\rm if}\ \frac{\| p_{\sigma}^{(t+1)} \|^2}{1 - (1-c_{\sigma})^{2(t+1)}} < \left( 2 + \frac{4}{d+1} \right) d, \\
0 & {\rm otherwise}.
\end{cases}
\end{align}

\noindent
{\bf Step 3. Updating of Distribution Parameters}\\
The distribution parameters are updated as follows~\cite{hansen2014principled}:
\begin{align}
    &m^{(t+1)} = m^{(t)} + c_m \sigma^{(t)} dy, \\
    &\sigma^{(t+1)} = \sigma^{(t)} \exp \left( \min \left(1, \frac{c_{\sigma}}{d_{\sigma}} \left( \frac{\| p_{\sigma}^{(t+1)} \|}{\mathbb{E}[\| \mathcal{N}(0, I) \|]} - 1 \right) \right) \right), \label{eq:stepsize} \\
    &\begin{multlined}[t]C^{(t+1)} = \left( 1 + (1-h_{\sigma}^{(t+1)}) c_1 c_c (2-c_c) \right) C^{(t)} \\
    +\underbrace{c_1 \left[ p_{c}^{(t+1)} \left( p_{c}^{(t+1)} \right)^{\top} - C^{(t)} \right]}_{\text{rank-one update}} + \underbrace{c_{\mu} \sum_{i=1}^{\mu} w_i \left[ y_{i:\lambda} y_{i:\lambda}^{\top} - C^{(t)} \right]}_{\text{rank-$\mu$ update}},\end{multlined}
\end{align}
where $\mathbb{E}[\| \mathcal{N}(0, I) \|] \approx \sqrt{d} \left( 1 - \frac{1}{4d} + \frac{1}{21 d^2} \right)$ is the expected Euclidean norm of the sample of a standard normal distribution;
$c_m$ is the learning rate for $m$, usually set to $1$;
$c_1$ and $c_{\mu}$ are the learning rates for the rank-one and -$\mu$ updates of $C$, respectively; and
$d_{\sigma}$ is a damping factor for the $\sigma$ adaptation.

\section{Learning Rate Adaptation Mechanism}
\label{sec:lra}

We consider the updating of the distribution parameters $\theta_m = m$ and $\theta_{\Sigma} = \vect (\Sigma)$, where $\vect$ is the vectorization operator and $\Sigma = \sigma^2 C$ in the case of the standard CMA-ES.
Let $\Delta_{m}^{(t)} = m^{(t+1)} - m^{(t)}$ and $\Delta_{\Sigma}^{(t)} = \vect (\Sigma^{(t+1)} - \Sigma^{(t)})$ be the original updates of $m$ and $\Sigma$, respectively.
We introduce the learning rate factors $\eta_m^{(t)}$ and $\eta_\Sigma^{(t)}$.
The modified updates are performed in form $\theta_m^{(t+1)} = \theta_m^{(t)} + \eta_m^{(t)} \Delta_m^{(t)}$ and $\theta_\Sigma^{(t+1)} = \theta_\Sigma^{(t)} + \eta_\Sigma^{(t)} \Delta_\Sigma^{(t)}$.
We then adapt $\etam^{(t)}$ and $\etasig^{(t)}$ separately.

\subsection{Main Concept}

We adapt the learning rate factor $\eta$ for a component $\theta$ (either $\theta_m = m$ or $\theta_{\Sigma} = \vect(\Sigma)$) of the distribution parameters based on the SNR of the update: 
\begin{equation}
    \mathrm{SNR} := \frac{\norm{\E[\Delta]}_{F}^2}{\Tr(F \Cov[\Delta])}
    = \frac{\norm{\E[\Delta]}_F^2}{\E[\norm{\Delta}^2_F] - \norm{\E[\Delta]}_F^2},
\end{equation}
where $F$ is the Fisher information matrix of the $\theta$ for which $\eta$ is to be updated, and $\norm{\Delta}_F = (\Delta^\T F \Delta)^{1/2}$ is the norm under $F$.
The choice of Fisher metric gives invariance property against parameterization of the probability distribution.
We attempt to adapt $\eta$ so that $\mathrm{SNR} = \alpha \eta$, where $\alpha > 0$ is a hyperparameter determining the target SNR.

The rationale behind this concept is as follows. 
Let us assume that $\eta$ is sufficiently small that the distribution parameters do not change significantly over $n$ iterations. 
That is, we assume $\theta^{(t + k)} \approx \theta^{(t)}$ for $k = 1, \dots, n$. 
Then, $\{\Delta^{(t + k)}\}_{k=0}^{n-1}$ are roughly considered to be independently and identically distributed. 
Hence, $n$ steps of the update read as
\begin{subequations}
\begin{align}
\theta^{(t+n)} 
&= \theta^{(t)} + \eta \sum_{k=0}^{n-1} \Delta^{(t+k)},
\\
&\approx \theta^{(t)} + \mathcal{D}\left( n\eta \E[\Delta], n\eta^2 \Cov[\Delta] \right),
\end{align}
\end{subequations}
where $\mathcal{D}(A, B)$ is the distribution with expectation $A$ and (co)variance $B$.
That is, by setting small $\eta$ and considering the results of $n = 1/\eta$ updates, we obtain an update that is more concentrated around the expected behavior than that expected for one update with $\eta = 1$.
The expected change of $\theta$ in $n = 1/\eta$ iterations measured by the squared Fisher norm, which approximates the Kullback-Leibler divergence between $\theta^{(t)}$ and $\theta^{(t+n)}$, is $\norm{\E[\Delta]}_F^2 + \eta\Tr(F \Cov[\Delta])$, where the former and latter terms come from the signal and noise, respectively. 
The SNR over $n$ iterations is $\frac{\norm{\E[\Delta]}_F^2}{\eta\Tr(F \Cov[\Delta])} = \frac{1}{\eta} \mathrm{SNR}$.
Therefore, maintaining $\mathrm{SNR} = \alpha \eta$ implies maintaining the SNR as $\alpha$ over $n = 1/\eta$ iterations, independently of $\eta$. 

\subsection{Signal-to-Noise Ratio Estimation}

We estimate $\norm{\E[\Delta]}^2$ and $\E[\norm{\Delta}^2]$ for each component ($m$ and $\Sigma$) with moving averages.
We let $\mvae^{(0)} = \bm{0}$ and $\mvav^{(0)} = 0$, and update them as
\begin{subequations}
\begin{align}
    \label{eq:mvae_update}
    \mvae^{(t+1)} &= (1 - \beta) \mvae^{(t)} + \beta \tilde\Delta^{(t)} ,\\
    \label{eq:mvav_update}
    \mvav^{(t+1)} &= (1 - \beta) \mvav^{(t)} + \beta \norm{ \tilde\Delta^{(t)} }_2^2 , 
\end{align}
\end{subequations}
respectively, where $\beta$ is a hyperparameter, $\tilde\Delta^{(t)}$ is the update at iteration $t$ in the local coordinate at which the $F$ at $\theta^{(t)}$ becomes the identity, and $\norm{\cdot}_2$ is the $\ell_2$-norm.
We then regard $\frac{2-\beta}{2 - 2\beta} \norm{\mvae}_2^2 - \frac{\beta}{2-2\beta} \mvav$ and $\mvav$ as the estimates of $\norm{\E[\Delta]}_2^2$ and $\E[\norm{\Delta}_2^2]$, respectively (See supplementary for the derivation). 

The rationale behind our estimators is as follows. 
Let us suppose that $\etam$ and $\etasig$ are sufficiently small for us to assume that the parameters $m$ and $\Sigma$ do not significantly change over $n$ iterations. 
Then, the $\tilde\Delta^{(t+i)}\ (i=0,.., n-1)$ are considered to be located on the same local coordinate and to be independently and identically distributed. 
Then, ignoring the $(1 - \beta)^n$ terms, we have
\begin{align}
    \label{eq:mvae_dist}
    \mvae^{(t+n)} 
    \sim \mathcal{D}\left( \E[\tilde\Delta], \frac{\beta}{2 - \beta} \Cov[\tilde\Delta]\right).
\end{align}
(See the supplementary material for the derivation.)
Therefore, we have $\E[\norm{\mvae}_2^2] \approx \norm{\E[\tilde\Delta]}_2^2 + \frac{\beta}{2-\beta}\Tr(\Cov[\tilde\Delta])$. 
Similarly, it is apparent that $\E[\mvav] \approx \E[\norm{\tilde\Delta}_2^2] = \norm{\E[\tilde\Delta]}_2^2 + \Tr(\Cov[\tilde\Delta])$.

The SNR is then estimated as
\begin{subequations}
\begin{align}
    \mathrm{SNR} &:= \frac{\norm{\E[\tilde\Delta]}^2}{\Tr(\Cov[\tilde\Delta])}
    = \frac{\norm{\E[\tilde\Delta]}^2}{\E[\norm{\tilde\Delta}^2] - \norm{\E[\tilde\Delta]}^2},
    \\
    &\approx \frac{\norm{\mvae}_2^2 - \frac{\beta}{2-\beta} \mvav}{\mvav - \norm{\mvae}_2^2}
    =: \widehat{\mathrm{SNR}}.
    \label{eq:snr_estimation}
\end{align}
\end{subequations}

\subsection{Learning Rate Factor Adaptation}

We attempt to adapt $\eta$ so that $\widehat{\mathrm{SNR}} = \alpha \eta$, where $\alpha > 0$ is the hyperparameter. 
This adaptation is expressed as  
\begin{equation}
    \label{eq:eta}
    \eta \leftarrow \eta \exp\left( \min(\gamma\eta, \beta) \Pi_{[-1,1]}\left( \frac{\widehat{\mathrm{SNR}}}{\alpha \eta} - 1 \right)\right),
\end{equation}
where $\Pi_{[-1, 1]}$ is the projection onto $[-1, 1]$ and $\gamma$ is a hyperparameter.
If $\widehat{\mathrm{SNR}} > \alpha \eta$, $\eta$ is increased, and vice versa. Because of these feedback mechanisms, $\widehat{\mathrm{SNR}} / (\alpha \eta)$ is expected to remain in the vicinity of $1$. 
In the above expression, the projection $\Pi_{[-1, 1]}$ is introduced to prevent a significant change of $\eta$ in one iteration. 
The damping factor $\min(\gamma\eta, \beta)$ is introduced for the following reasons. 
First, the factor $\beta$ is introduced to allow for the effect of the change of the previous $\eta$ to appear in $\widehat{\mathrm{SNR}}$. 
Second, the factor $\gamma\eta$ is introduced to  prevent $\eta$ from changing to a greater extent than the factor of $\exp(\gamma)$ or $\exp(-\gamma)$ in $1/\eta$ iterations. 
Following updating of $\eta$ by Eq.~(\ref{eq:eta}), we set the upper bound to $1$ using $\eta \leftarrow \min(\eta, 1)$, to prevent unstable behavior.
Allowing $\eta$ to take values exceeding $1$ would accelerate the optimization; however, this aspect is not considered in the present study, as the aim is to safely solve difficult problems.

\subsection{ Local Coordinate-System Definition}

We define the local coordinate system such that the Fisher information matrices, $F_m$ and $F_{\Sigma}$, corresponding to each component of the distribution parameters, $m$ and $\Sigma$, respectively, are the identity matrices.
It is well-known that $F_m = \Sigma^{-1}$ and $F_{\Sigma} = 2^{-1} \Sigma^{-1} \otimes \Sigma^{-1}$. 
Their square roots are $\sqrt{F_m} = \sqrt{\Sigma}^{-1}$ and $\sqrt{F_{\Sigma}} = 2^{-\frac{1}{2}} \sqrt{\Sigma}^{-1} \otimes \sqrt{\Sigma}^{-1}$.
Therefore, we define
\begin{subequations}
\begin{align}
    \tilde{\Delta}_m &= \sqrt{\Sigma}^{-1} \Delta_m, \\
    \tilde{\Delta}_{\Sigma} &= 2^{-\frac{1}{2}} \vect(\sqrt{\Sigma}^{-1} \vect^{-1}(\Delta_{\Sigma}) \sqrt{\Sigma}^{-1}).
\end{align}
\end{subequations}

\subsection{Covariance Matrix Decomposition}

Following computation of the updated covariance matrix $\Sigma^{(t+1)} = \Sigma^{(t)} + \eta_{\Sigma}^{(t)} \vect^{-1} (\Delta_{\Sigma}^{(t)})$, we must split the matrix into components $\sigma$ and $C$. 
For this purpose, we adopt the following, simple strategy:
\begin{subequations}
\begin{align}
    \sigma^{(t+1)} &= \det(\Sigma^{(t+1)})^{\frac{1}{2d}}, \\
    C^{(t+1)} &= (\sigma^{(t+1)})^{-2} \Sigma^{(t+1)}.   
\end{align}
\end{subequations}

\subsection{Step-Size Correction}

When the learning rate for the $m$ update, i.e., $\eta_m$, is updated, the appropriate $\sigma$ changes. Through a quality gain analysis in which the expected improvement of the $f$ value in one step was analyzed, a previous study \cite{qualitygainwres} demonstrated that the optimal $\sigma$ is proportional to $1/\eta_m$ for infinite-dimensional convex quadratic functions. To maintain the optimal $\sigma$ while $\eta_m$ varies, we correct $\sigma$ after each $\eta_m$ update as follows:
\begin{equation}
\sigma^{(t+1)} \leftarrow \frac{\eta_m^{(t)}}{\eta_{m}^{(t+1)}} \sigma^{(t+1)} .
\end{equation}

\subsection{Overall Procedure}

Algorithm~\ref{alg:cma_learningrate} shows the overall LRA-CMA-ES procedure.
In line 2 of Algorithm~\ref{alg:cma_learningrate}, {\sf{CMA($\cdot$)}} receives the old parameters $m^{(t)}, \sigma^{(t)},$ and $C^{(t)}$ and outputs new parameters $m^{(t+1)}, \sigma^{(t+1)},$ and $C^{(t+1)}$ by running Steps 1--3 described in Sec.~\ref{sec:cma}.
Note that the internal parameters, such as the evolution paths $p_{\sigma}$ and $p_c$, are updated and stored in {\sf{CMA($\cdot$)}}. They are now omitted for simplicity.
The subscript $\cdot_{\{ m,\Sigma \}}$ (e.g., as in $\eta_{\{ m,\Sigma \}}$) indicates that there are parameters for $m$ and $\Sigma$, respectively.
For example, $\mvae^{(t+1)}_{\{ m,\Sigma \}} \gets (1 - \beta_{\{ m,\Sigma \}}) \mvae^{(t)}_{\{ m,\Sigma \}} + \beta_{\{ m,\Sigma \}} \tilde\Delta^{(t)}_{\{ m,\Sigma \}}$ is an abbreviation for the following two update equations: $\mvae^{(t+1)}_{m} \gets (1 - \beta_{m}) \mvae^{(t)}_{m} + \beta_{m} \tilde\Delta^{(t)}_{m}$ and $\mvae^{(t+1)}_{\Sigma} \gets (1 - \beta_{\Sigma}) \mvae^{(t)}_{\Sigma} + \beta_{\Sigma} \tilde\Delta^{(t)}_{\Sigma}$.

\begin{algorithm}
\caption{LRA-CMA-ES}
\label{alg:cma_learningrate}
\begin{algorithmic}[1]
\Require $m^{(0)} \in \mathbb{R}^d, \sigma^{(0)} \in \mathbb{R}_{>0}, \lambda \in \mathbb{N}, \alpha, \beta_{\{ m,\Sigma \}}, \gamma \in \mathbb{R}$
\Set $t = 0, C^{(0)}=I, \eta_{\{m,\Sigma\}}^{(0)} = 1, \mvae^{(0)} = \bm{0}, \mvav^{(0)} = 0$
\While{stopping criterion not met}
    \State $m^{(t+1)}, \sigma^{(t+1)}, C^{(t+1)} \gets {\rm CMA}(m^{(t)}, \sigma^{(t)}, C^{(t)})$
    \State {\sf \//\// calculate parameter one-step differences}
    \State $\Delta_m^{(t)} \gets m^{(t+1)} - m^{(t)}$
    \State $\Sigma^{(t+1)} \gets \left( \sigma^{(t+1)} \right)^2 C^{(t+1)}$
    \State $\Delta_{\Sigma}^{(t)} \gets {\rm vec}\left( \Sigma^{(t+1)} - \Sigma^{(t)} \right)$
    \State {\sf \//\// local coordinate}
    \State $\tilde{\Delta}_m^{(t)} \gets \sqrt{\Sigma^{(t)}}^{-1} \Delta_m^{(t)}$
    \State $\tilde{\Delta}_{\Sigma}^{(t)} \gets 2^{-1/2} {\rm vec}\left(\sqrt{\Sigma^{(t)}}^{-1} {\rm vec}^{-1}\left(\Delta_{\Sigma}^{(t)}\right) \sqrt{\Sigma^{(t)}}^{-1} \right)$
    \State {\sf \//\// update evolution paths and estimate SNR}
    \State $\mvae^{(t+1)}_{\{ m,\Sigma \}} \gets (1 - \beta_{\{ m,\Sigma \}}) \mvae^{(t)}_{\{ m,\Sigma \}} + \beta_{\{ m,\Sigma \}} \tilde\Delta^{(t)}_{\{ m,\Sigma \}}$
    \State $\mvav^{(t+1)}_{\{ m,\Sigma \}} \gets (1 - \beta_{\{ m,\Sigma \}}) \mvav^{(t)}_{\{ m,\Sigma \}} + \beta_{\{ m,\Sigma \}} \norm{ \tilde\Delta^{(t)}_{\{ m,\Sigma \}} }_2^2$
    \State $\widehat{\mathrm{SNR}}_{\{ m,\Sigma \}} \gets \frac{\norm{\mvae^{(t+1)}_{\{ m,\Sigma \}} }_2^2 - \frac{\beta_{\{ m,\Sigma \}} }{2-\beta_{\{ m,\Sigma \}} } \mvav^{(t+1)}_{\{ m,\Sigma \}} }{\mvav^{(t+1)}_{\{ m,\Sigma \}} - \norm{\mvae^{(t+1)}_{\{ m,\Sigma \}}}_2^2}$
    \State {\sf \//\// update learning rates}
    \State $\eta_{\{ m,\Sigma \}}^{(t+1)} \leftarrow \eta_{\{ m,\Sigma \}}^{(t)}$
    \Statex $\cdot \exp\left( \min(\gamma \eta_{\{ m,\Sigma \}}^{(t)}, \beta_{\{ m,\Sigma \}}) \Pi_{[-1,1]}\left( \frac{\widehat{\mathrm{SNR}}_{\{ m,\Sigma \}} }{\alpha \eta_{\{ m,\Sigma \}} } - 1 \right)\right)$
    \State $\eta_{\{ m,\Sigma \}}^{(t+1)} \leftarrow \min(\eta_{\{ m,\Sigma \}}^{(t+1)}, 1)$
    \State {\sf \//\// update parameters with adaptive learning rates}
    \State $m^{(t+1)} \gets m^{(t)} + \eta_m^{(t+1)} \Delta_m^{(t)}$
    \State $\Sigma^{(t+1)} \gets \Sigma^{(t)} + \eta_{\Sigma}^{(t+1)} \vect^{-1} (\Delta_{\Sigma}^{(t)})$
    \State {\sf \//\// decompose $\Sigma$ to $\sigma$ and $C$}
    \State $\sigma^{(t+1)} \gets \det(\Sigma^{(t+1)})^{\frac{1}{2d}}$, $C^{(t+1)} \gets (\sigma^{(t+1)})^{-2} \Sigma^{(t+1)}$
    \State {\sf \//\// $\sigma$ correction}
    \State $\sigma^{(t+1)} \gets \sigma^{(t+1)} (\eta_{m}^{(t)} / \eta_{m}^{(t+1)})$
    \State $t \gets t + 1$
\EndWhile
\end{algorithmic}
\end{algorithm}

\section{Experiments}
\label{sec:exp}
In this study, we performed various experiments to investigate the following research questions (RQs):

\begin{itemize}
    \item [\textbf{RQ1.}] Does the $\eta$ adaptation in the LRA-CMA-ES behave appropriately in accordance with the problem structure?
    \item [\textbf{RQ2.}] Can the LRA-CMA-ES solve multimodal and noisy problems even though a default $\lambda$ is used? How does its efficiency compare to the CMA-ES with fixed $\eta$?
    \item [\textbf{RQ3.}] How does the performance change when the LRA-CMA-ES hyperparameters are changed?
\end{itemize}

This section is organized as follows: the experiment setups are described in Sec.\ref{sec:setup}; in Sec.\ref{sec:lrbahavior}, the $\eta$ adaptation in the LRA-CMA-ES for noiseless and noisy problems (RQ1) is demonstrated; in Sec.~\ref{sec:fixed_vs_adapt}, we compare the LRA-CMA-ES with the CMA-ES with fixed $\eta$ values (RQ2); finally, in Sec.~\ref{sec:exp_hyperparam}, our investigation of the effect of the LRA-CMA-ES hyperparameters (RQ3) is reported.
Our code is available at \textcolor{blue}{https://github.com/nomuramasahir0/cma-learning-rate-adaptation}.

\subsection{Experiment Setups}
\label{sec:setup}
The benchmark problem definitions and initial distributions are listed in Table~\ref{tab:benchmark}.
In each case (except for the Rosenbrock function), the known global optimal solution is located at $x = 0$. For the Rosenbrock function, the global optimal solution is $x = 1$.
Note that, although the Rosenbrock function has local minima, in our setting, it could be regarded as an almost unimodal problem.
Similar to \cite{hansen2004evaluating}, we imposed additional bounds
for the Ackley function.
In the noisy problems, we considered the additive Gaussian noise $\epsilon \sim \mathcal{N}(0, \sigma_n^2)$ with variance $\sigma_n^2$.

In all experiments, we used the default $\lambda = 4 + \lfloor 3 \ln d \rfloor$.
As regards the LRA-CMA-ES hyperparameters, we set $\alpha = 1.4$, $\beta_m = 0.1$, $\beta_{\Sigma} = 0.03$, and $\gamma = 0.1$ based on preliminary experiments.
As noted above, Sec.~\ref{sec:exp_hyperparam} reports our analysis of the sensitivity of these hyperparameters.
The other internal parameters of the CMA-ES were set to the values recommended in \cite{hansen2014principled}.

\begin{table*}[t]
  \centering
  \caption{Definitions of benchmark problems and initial distributions used in experiments.}
  \label{tab:benchmark}
  \begin{tabular}{l|l}
    \bottomrule
    Definitions & Initial Distributions \\
    \hline \hline
    $f_{\rm Sphere}(x) = \sum_{i=1}^{d} x_i^2$ & $m^{(0)} = [3,\ldots,3], \sigma^{(0)} = 2$ \\
    $f_{\rm Ellipsoid}(x) = \sum_{i=1}^{d} (1000^{\frac{i-1}{d-1}}x_i)^2$ & $m^{(0)} = [3,\ldots,3], \sigma^{(0)} = 2$ \\
     $f_{\rm Rosenbrock}(x) = \sum_{i=1}^{d-1} (100 (x_{i+1} - x_i^2)^2 + (x_i - 1)^2)$ & $m^{(0)} = [0,\ldots,0], \sigma^{(0)} = 0.1$ \\
     $f_{\rm Ackley}(x) = 20 - 20 \cdot \exp (-0.2 \sqrt{\frac{1}{d} \sum_{i=1}^d x_i^2}) + e - \exp (\frac{1}{d} \sum_{i=1}^d \cos(2\pi x_i))$ & $m^{(0)} = [15.5,\ldots,15.5], \sigma^{(0)} = 14.5$ \\
     $f_{\rm Schaffer}(x) = \sum_{i=1}^{d-1} (x_i^2 + x_{i+1}^2)^{0.25} \cdot [\sin^2(50\cdot (x_i^2 + x_{i+1}^2)^{0.1}) + 1]$ & $m^{(0)} = [55,\ldots,55], \sigma^{(0)} = 45$ \\
     $f_{\rm Rastrigin}(x) = 10d + \sum_{i=1}^d (x_i^2 - 10 \cos(2\pi x_i))$ & $m^{(0)} = [3,\ldots,3], \sigma^{(0)} = 2$ \\
     $f_{\rm Bohachevsky}(x) = \sum_{i=1}^{d-1} ( x_i^2 + 2x_{i+1}^2 - 0.3 \cos(3\pi x_i) - 0.4\cos(4\pi x_{i+1}) + 0.7 )$ & $m^{(0)} = [8,\ldots,8], \sigma^{(0)} = 7$ \\
     $f_{\rm Griewank}(x) = \frac{1}{4000} \sum_{i=1}^d x_i^2 - \Pi_{i=1}^d \cos (x_i / \sqrt{i}) + 1$ & $m^{(0)} = [305,\ldots,305], \sigma^{(0)} = 295$ \\
    \bottomrule
  \end{tabular}
\end{table*}

\begin{figure*}[tb]
  \centering
  \includegraphics[width=0.964\hsize,trim=5 5 5 5,clip]{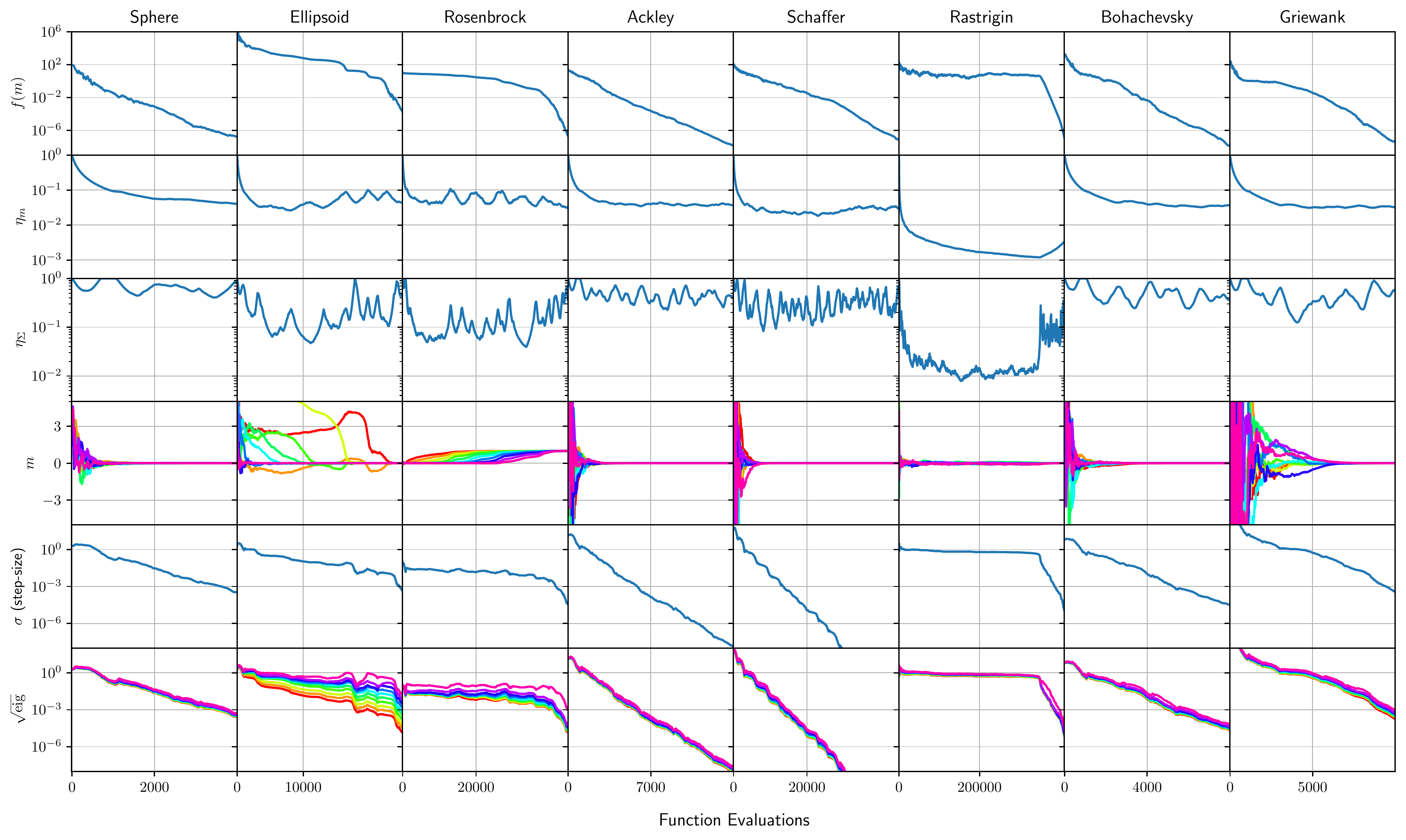}
  \caption{Typical LRA-CMA-ES behavior on 10-dimensional (10-D) noiseless problems. The coordinates of $m$ and the square roots of the eigenvalues of $\sigma^2 C$ (denoted by $\sqrt{\text{eig}}$) are shown in different colors.}
  \label{fig:adaptlr_behavior}
\end{figure*}

\begin{figure}[tb]
  \centering
  \includegraphics[width=0.964\hsize,trim=5 5 5 5,clip]{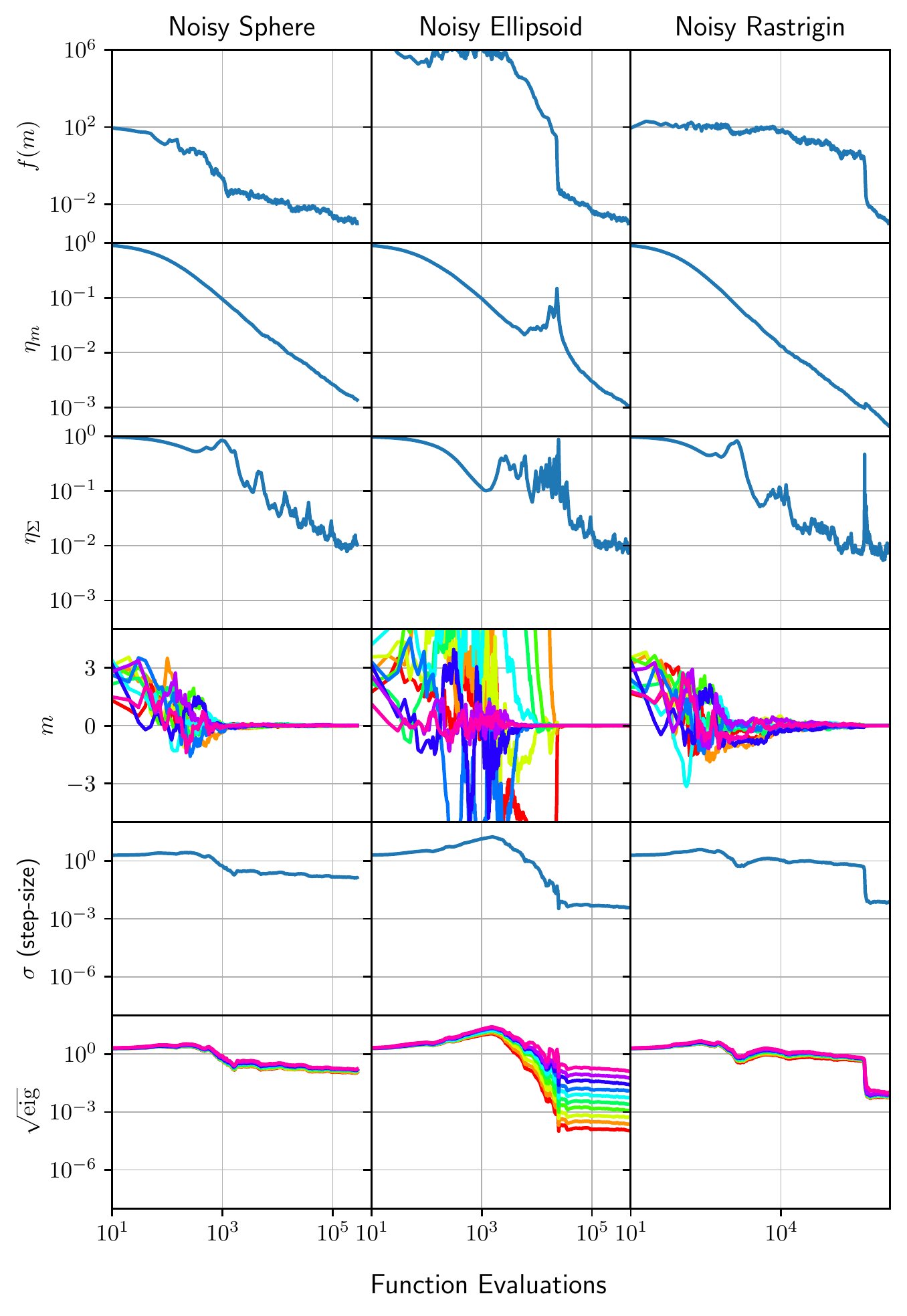}
  \caption{Typical LRA-CMA-ES behavior on 10-D noisy problems. The noise variance $\sigma_n^2$ was set to $1$.}
  \label{fig:adaptlr_behavior_noisy_loglinear}
\end{figure}

\subsection{Learning Rate Behavior}
\label{sec:lrbahavior}
Figure~\ref{fig:adaptlr_behavior} shows typical LRA-CMA-ES behavior on noiseless problems.
It is apparent that $\eta_{\Sigma}$ retained relatively high values for the Sphere function.
However, $\eta_{\Sigma}$ decreased significantly for the Ellipsoid and Rosenbrock functions.
We believe that this behavior is undesirable, because the default $\eta$ already works well on these unimodal problems.
We can increase $\eta$ by changing the hyperparameters of the proposed $\eta$ adaptation; however, this change may be detrimental as regards applications to multimodal problems.

It is apparent that $\eta_m$ was slightly smaller on multimodal problems than on unimodal problems.
For the Rastrigin function in particular, $\eta_m$ and $\eta_{\Sigma}$ clearly decreased at the beginning of the optimization; this reflects the difficulty of optimization for multimodal problems.
Subsequently, $\eta$ increased, as the optimization became as easy as that for a unimodal problem. 
This behavior demonstrates that the LRA-CMA-ES can adapt $\eta$ according to the difficulty of the search situation.

Figure~\ref{fig:adaptlr_behavior_noisy_loglinear} shows typical $\eta$ adaptation behavior on noisy problems.
The noise had an essentially negligible effect in the early stage and, thus, the $\eta$ behavior on the noisy problems was similar to that on the noiseless problems.
However, as the optimization proceeded and the function value approached the same scale as the noise value, the noise began to have an critical effect.
In response, $\eta$ decreased. Through this adaptation, the SNR remained constant.
Notably, similar behavior was obtained for the Noisy Rastrigin function, which featured both noise and multimodality.

\subsection{Learning Rate Adaptation Effect}
\label{sec:fixed_vs_adapt}

\begin{figure*}[tb]
  \centering
  \includegraphics[width=0.964\hsize,trim=5 5 5 5,clip]{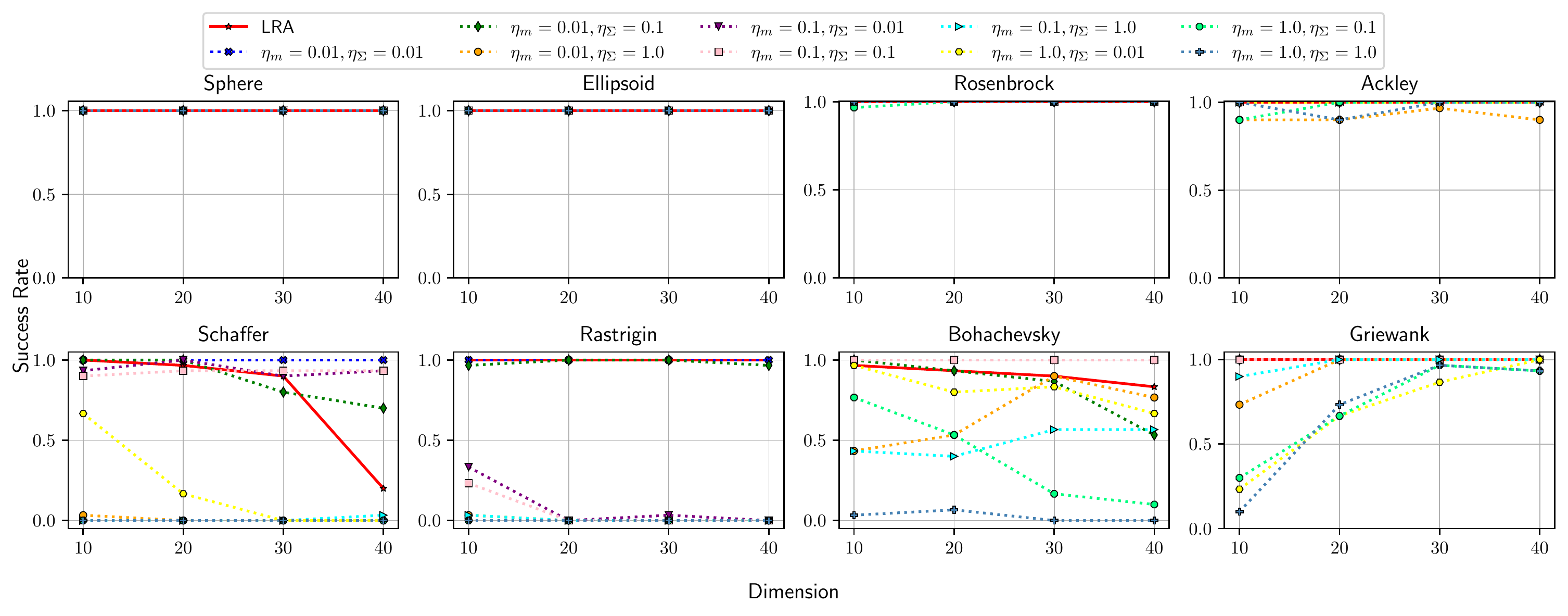}
  \caption{Success rate versus dimension (noiseless problems).}
  \label{fig:compare_succrate}
\end{figure*}

\begin{figure*}[tb]
  \centering
  \includegraphics[width=0.964\hsize,trim=5 5 5 5,clip]{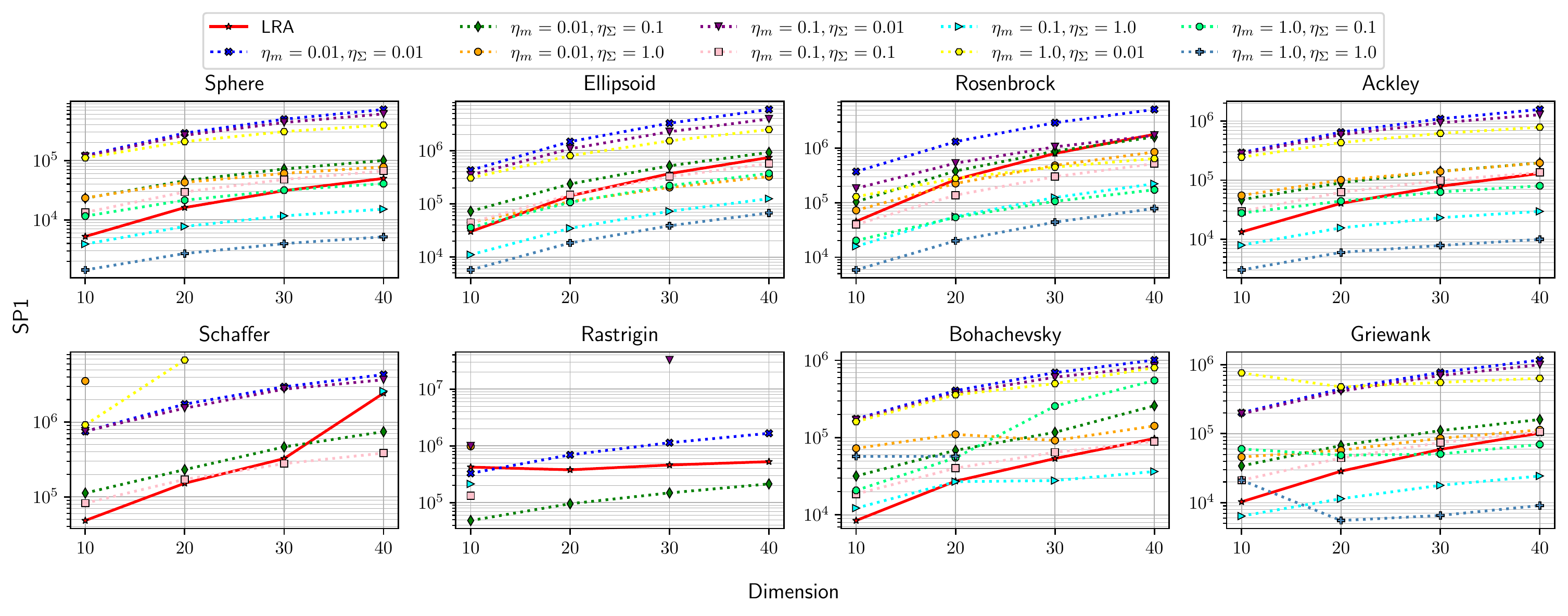}
  \caption{SP1 versus dimension (noiseless problems).}
  \label{fig:compare_sp1}
\end{figure*}

Figures~\ref{fig:compare_succrate} and ~\ref{fig:compare_sp1} show the performance of the LRA-CMA-ES and that of the CMA-ES with a fixed learning rate $(\eta_m, \eta_{\Sigma} \in \{10^0, 10^{-1}, 10^{-2} \})$ on the noiseless problems.
Note that the CMA-ES with $\eta_m = 1.0$ and $\eta_{\Sigma} = 1.0$ was the CMA-ES with the default $\eta$ (i.e., the original CMA-ES).
Each trial was considered a success if $f(m)$ reached the target value $10^{-8}$ before $10^7$ function evaluations, or before a numerical error occurred as a result of excessively small $\sigma$.
In addition to the success rate, we also employed the SP1~\cite{auger2005restart}, i.e., the average number of evaluations among successful trials until the target value was reached divided by the success rate.
We performed 30 trials for each setting.

To compare the performance of these strategies on the selected noisy problems, we employed the empirical cumulative density function (ECDF) used in COCO, which is a platform for comparing continuous optimizers in a black-box setting~\cite{hansen2021coco}.
Using $N_{\rm target}$ target values, we recorded the number of evaluations until the (noiseless) $f(m)$ reached each target value for the first time, with the maximum function evaluation set to $10^8$.
We collected data by running $N_{\rm trial}$ independent trials, and obtained a total of $N_{\rm target} \cdot N_{\rm trial}$ targets for each problem.
We set the target values to $10^{6 - 9(i-1)/(N_{\rm target} - 1)}$ for $i = 1, \ldots, N_{\rm target}$, with $N_{\rm target} = 30$. By running $N_{\rm trial} = 20$ trials, we had 600 targets for each problem.
Figure~\ref{fig:ecdf} shows the percentage of target values achieved for each number of evaluations.

\subsubsection{Noiseless Problems}
We compared the success rates of the LRA-CMA-ES and the CMA-ES with fixed $\eta$ values, as shown in Figure~\ref{fig:compare_succrate}.
For the multimodal problems, the CMA-ES with high $\eta$ often failed to reach the optimum.
However, the CMA-ES with small $\eta$ had a high success rate, demonstrating the clear performance dependence on $\eta$.
In contrast, the LRA-CMA-ES had a relatively good success rate even though no $\eta$ tuning was required.
It is noteworthy that the LRA-CMA-ES succeeded in all trials on the Rastrigin function, even though the default sample size (e.g., $\lambda = 15$ for $d=40$) was used and $\eta$ was not tuned in advance.

The LRA-CMA-ES performance degraded on the Schaffer function with $d=40$.
From the results indicating that the CMA-ES with an appropriately tuned, small $\eta$ achieved a relatively high success rate, the LRA-CMA-ES result may have been obtained because $\eta$ was not appropriately adapted in that case.
We leave a further investigation for future work.

Figure~\ref{fig:compare_sp1} shows the SP1 results for the LRA-CMA-ES and the CMA-ES with fixed $\eta$ values.
The CMA-ES with the default $\eta$ ($\eta_m = 1.0, \eta_{\Sigma} = 1.0$) outperformed the other methods on the unimodal problems; however, this performance degraded significantly on the multimodal problems as a result of optimization failure.
In contrast, the CMA-ES with small $\eta$ sometimes exhibited good performance on such multimodal problems, but did not exhibit efficiency when applied to the unimodal and relatively easy multimodal problems.
Therefore, for the CMA-ES with fixed $\eta$, a clear trade-off in efficiency exists depending on the $\eta$ setting.
In contrast, the LRA-CMA-ES exhibited stable and relatively good performance when applied to unimodal and multimodal problems.
Again, no tuning of $\eta$ was performed, which is extremely expensive in practice.
There is scope for improvement of the LRA-CMA-ES performance on unimodal problems; however, the current sub-par performance can be somewhat mitigated by changing the hyperparameters.
Again, we discuss the impact of the hyperparameters in Sec.~\ref{sec:exp_hyperparam}.

\subsubsection{Noisy Problems}

\begin{figure*}[tb]
  \centering
  \includegraphics[width=0.964\hsize,trim=5 5 5 5,clip]{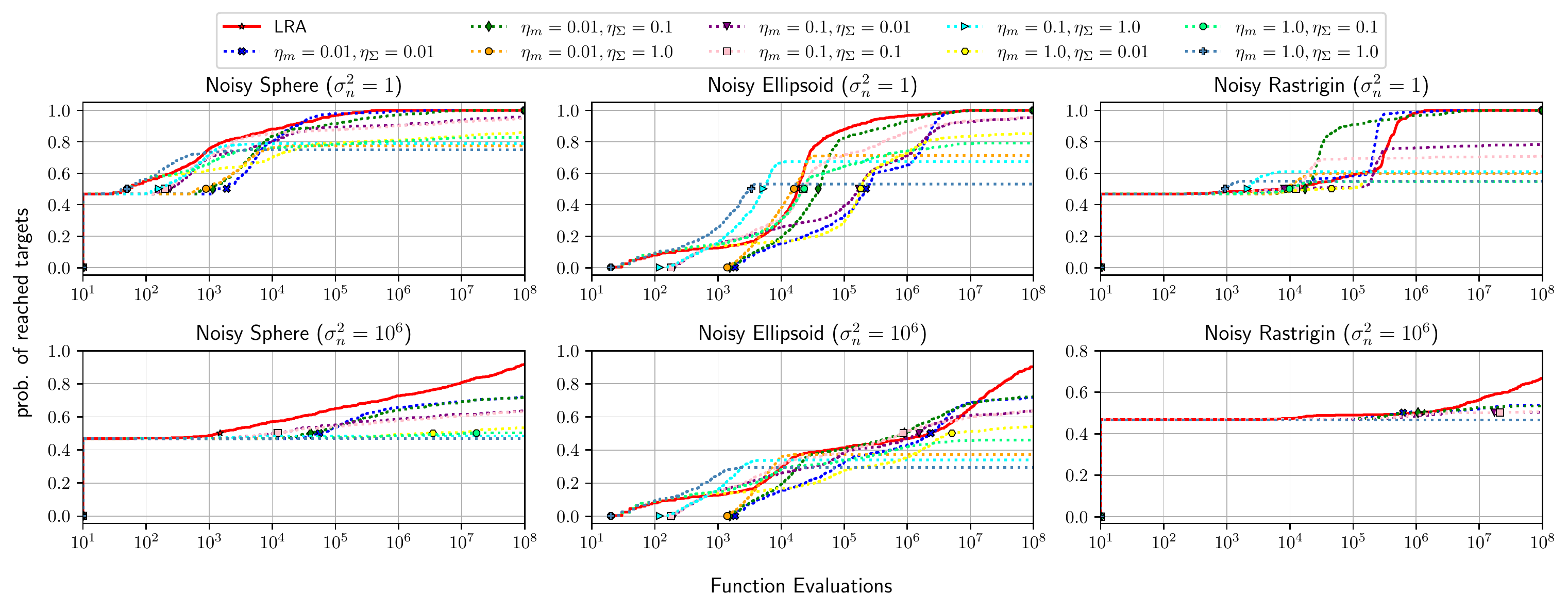}
  \caption{Empirical cumulative density function on 10-D noisy problems, with $\sigma_n^2$ set to $1$ or $10^6$.}
  \label{fig:ecdf}
\end{figure*}

Figure~\ref{fig:ecdf} shows the ECDF results for the LRA-CMA-ES and for the CMA-ES with fixed $\eta$ values.
We considered two noise strengths: weak and strong, i.e., $\sigma_n^2 = 1$ and $10^6$, respectively.

For the weak noise settings, the CMA-ES with small $\eta$ reached all target values. In contrast, the CMA-ES with large $\eta$ failed to approach the global optimum and yielded a sub-optimal solution.
The LRA-CMA-ES achieved similar performance to the CMA-ES with small $\eta$ without tuning of $\eta$. For the strong noise settings, even the CMA-ES with small $\eta$ stopped improving the $f$ value before the global optimum was reached.
In contrast, the LRA-CMA-ES continued to improve the $f$ value.
Notably, the results for the Noisy Rastrigin function suggest that the LRA-CMA-ES can simultaneously handle both noise and multimodality.

\subsection{Effects of Hyperparameters}
\label{sec:exp_hyperparam}

\begin{figure*}[tb]
  \centering
  \includegraphics[width=0.964\hsize,trim=5 10 5 25,clip]{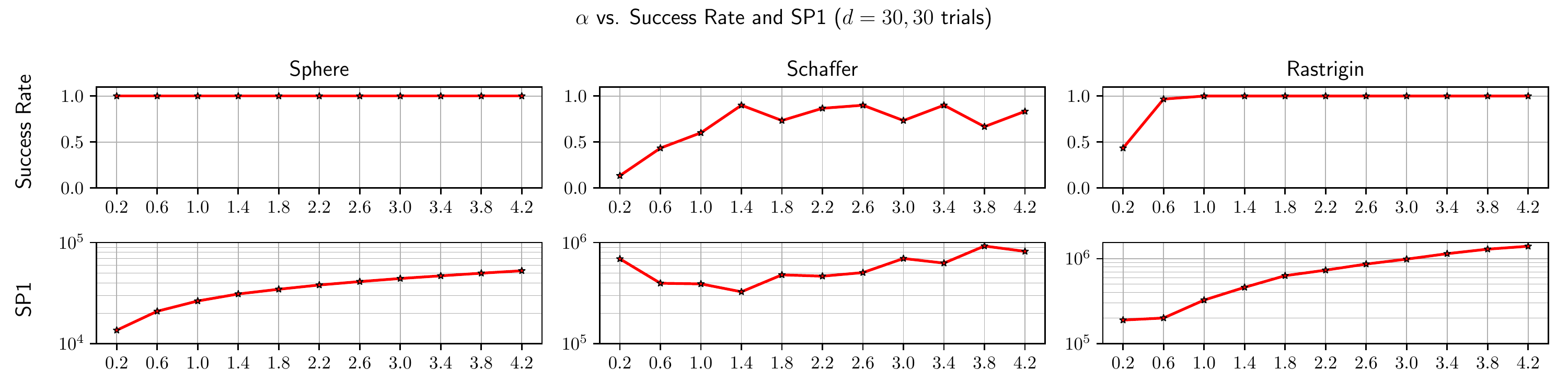}
  \caption{Success rate and SP1 versus hyperparameter $\alpha$ on 30-D noiseless problems (30 trials).}
  \label{fig:alpha_srsp1_d=30}
\end{figure*}

\begin{figure*}[tb]
  \centering
  \includegraphics[width=0.964\hsize,trim=5 10 5 25,clip]{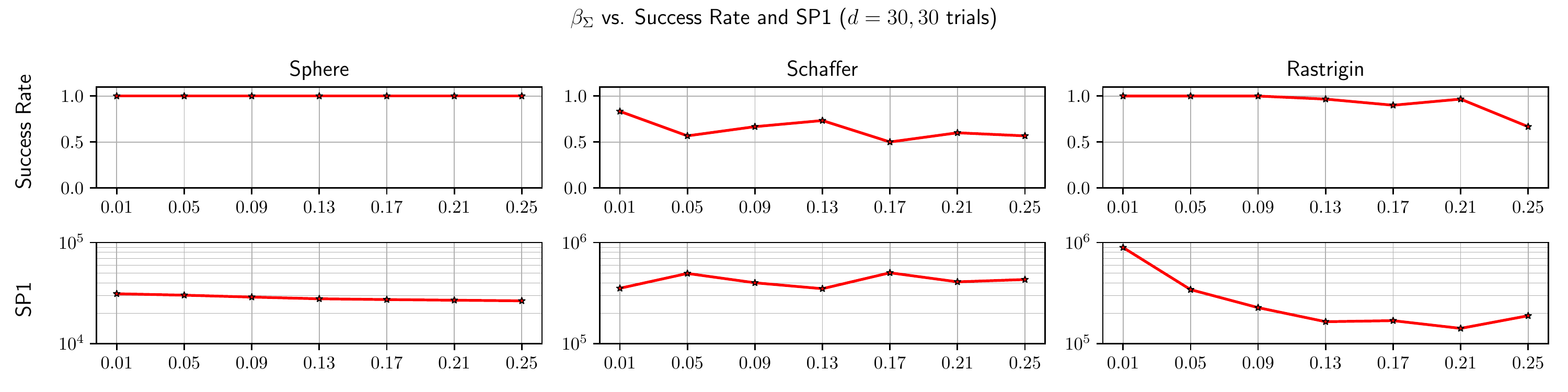}
  \caption{Success rate and SP1 versus hyperparameter $\beta_{\Sigma}$ on 30-D noiseless problems (30 trials).}
  \label{fig:beta_Sigma_srsp1_d=30}
\end{figure*}

Figure~\ref{fig:alpha_srsp1_d=30} shows the success rate and SP1 values with respect to $\alpha$ on the 30-dimensional (30-D) noiseless Sphere, Schaffer, and Rastrigin functions.
For the Sphere function, it is apparent that smaller SP1 could be achieved for a smaller $\alpha$ value.
However, excessively small $\alpha$ yielded optimization failures for the multimodal problems.
The current setting of $\alpha = 1.4$ seems reasonable, but further investigation is necessary.

Figure~\ref{fig:beta_Sigma_srsp1_d=30} shows the success rate and SP1 values with respect to $\beta_{\Sigma}$.
Clearly, excessively large $\beta_{\Sigma}$ caused optimization failures for the multimodal problems.
However, setting an excessively small $\beta_{\Sigma}$ yielded slow convergence for the Rastrigin function.
An additional result ($\beta_{\Sigma} \in \{ 0.01, 0.02, ..., 0.05 \}$) is presented in the supplementary material.

We also conducted similar experiments on the hyperparameters $\beta_m$ and $\gamma$, to confirm their effects.
These hyperparameters appear to have had a mild impact on the overall performance compared to $\alpha$ and $\beta_{\Sigma}$.
(These results are also presented in the supplementary material.)

\section{Conclusion}
\label{sec:conclusion}
In this study, we developed a new learning rate adaptation mechanism with the aim of solving multimodal and noisy problems via the CMA-ES with a default population size.
The basic concept of the resulting algorithm, LRA-CMA-ES, is to adapt the learning rate so that the SNR can be kept constant.
Experiments involving noiseless multimodal problems revealed that the proposed LRA-CMA-ES can adapt the learning rate appropriately depending on the search situation, and that it works well without tuning of the learning rate.
In noisy experiments, the LRA-CMA-ES continued to yield improved solution quality. This performance was observed even for strong noise settings, which yielded problems that the CMA-ES with a fixed learning rate failed to solve.
In conclusion, the LRA-CMA-ES with default population size facilitates solution of multimodal and noisy problems to some extent, without the need for tuning of the learning rate.

However, we acknowledge limitations of the proposed LRA-CMA-ES, which will direct our future work.
First, we observed that the LRA-CMA-ES experienced many failures for the 40-D Schaffer function, although the CMA-ES with an appropriately small learning rate succeeded with high probability.
We believe that detailed analysis of the behavior of the SNR adaptation mechanism is important to clarify the reasons for this failure.
On a related note, our understanding of the appropriate hyperparameter setting in the proposed learning rate adaptation mechanism remains limited.
Our experiments revealed that the hyperparameter setting affects the trade-off between stability and convergence speed.
Through experiment, we identified the hyperparameters that perform relatively well on noiseless and noisy problems; however, development of better configuration methods is important.
For example, the constant value $\mathcal{O}(1)$ is used for the cumulation factors $\beta_m$ and $\beta_{\Sigma}$, but it may be more reasonable to have these factors depend on the parameter degrees of freedom, i.e., $\beta_m = \mathcal{O}(1/d)$ and $\beta_{\Sigma} = \mathcal{O}(1/d^2)$.
A deeper understanding of the effects of the hyperparameters including $\alpha$ is critical to increasing the reliability of the proposed LRA-CMA-ES.
Finally, a performance comparison of learning rate adaptation against population size adaptation is another interesting direction for future research.

\begin{acks}
This study was partially supported by JSPS KAKENHI, grant number 19H04179.
\end{acks}

\bibliographystyle{ACM-Reference-Format}
\balance
\bibliography{ref}

\appendix

\section{Derivation for Section 3.2}
\label{sec:app_snr}

\subsection{Derivation of Eq.~(12)}
This section presents the detailed derivation of Eq.~(12).
By ignoring $(1-\beta)^n$, $\mvae^{(t+n)}$ can be approximately calculated as follows:
\small
\begin{align*}
    \mvae^{(t+n)} &= (1 - \beta) \mvae^{(t+n-1)} + \beta \tilde\Delta^{(t+n-1)} \\
    &= (1-\beta) \left\{ (1-\beta) \mvae^{(t+n-2)} + \beta \tilde\Delta^{(t+n-2)} \right\} + \beta \tilde\Delta^{(t+n-1)} \\
    &= ...\\
    &= (1-\beta)^{n} \mvae^{(t)} + \sum_{i=0}^{n-1} (1-\beta)^i \beta \tilde\Delta^{(t+n-1-i)} \\
    &\approx \sum_{i=0}^{n-1} (1-\beta)^i \beta \tilde\Delta^{(t+n-1-i)}.
\end{align*}
\normalsize
Here, we assume the $\tilde\Delta^{(\cdot)}$ are uncorrelated with each other; this corresponds to the scenario where $\eta$ is sufficiently small.
In this case, we can ignore the dependence of $t$, i.e., $\E[\tilde\Delta^{(t+n-1-i)}] =: \E[\tilde\Delta]$.
Thus,
\small
\begin{align*}
    \E[\mvae^{(t+n)}] = \sum_{i=0}^{n-1} (1-\beta)^i \beta \E[\tilde\Delta].
\end{align*}
\normalsize
Here,
\small
\begin{align*}
     \sum_{i=0}^{n-1} (1-\beta)^i &= \frac{1 \cdot \{ 1-(1-\beta)^{n} \} }{1-(1-\beta)} = \frac{1-(1-\beta)^{n}}{\beta}.
\end{align*}
\normalsize
Then, by ignoring $(1-\beta)^{n}$, we can approximate $\E[\mvae^{(t+n)}]$ as follows:
\small
\begin{align*}
    \E[\mvae^{(t+n)}] = [1-(1-\beta)^{n}] \E[\tilde\Delta] \approx \E[\tilde\Delta].
\end{align*}
\normalsize
Next, we consider the covariance $\Cov[\mvae^{(t+n)}]$:
\small
\begin{align*}
    \Cov[\mvae^{(t+n)}] = \E[\mvae^{(t+n)} (\mvae^{(t+n)})^{\top}] - \E[[\mvae^{(t+n)}] ([\mvae^{(t+n)}])^{\top}.
\end{align*}
\normalsize
First, we find the exact expression of $\mvae^{(t+n)} (\mvae^{(t+n)})^{\top}$:
\small
\begin{align*}
    &\mvae^{(t+n)} (\mvae^{(t+n)})^{\top} = \beta^2 \sum_{i=0}^{n-1} (1-\beta)^{2i} \tilde\Delta^{(t+n-1-i)} (\tilde\Delta^{(t+n-1-i)})^{\top} \\
    &\quad + 2\beta^2 \sum_{i,j=0: i\neq j}^{n-1} (1-\beta)^i (1-\beta)^j \tilde\Delta^{(t+n-1-i)} (\tilde\Delta^{(t+n-1-i)})^{\top}.
\end{align*}
\normalsize
Note that, for $i,j \in \{0, \cdots n-1 \} (i \neq j)$, $\E[\tilde\Delta^{(t+n-1-i)} (\tilde\Delta^{(t+n-1-j)})^{\top}]$ $= \E[\tilde\Delta] (\E[\tilde\Delta])^{\top}$, as we assume that they are uncorrelated.
For $i \in \{0, \cdots n-1 \}$, $\E[\tilde\Delta^{(t+n-1-i)} (\tilde\Delta^{(t+n-1-i)})^{\top}] = \E[\tilde\Delta] (\E[\tilde\Delta])^{\top} + \Cov[\tilde\Delta]$.
Thus,
\small
\begin{align*}
    \E[\mvae^{(t+n)}& (\mvae^{(t+n)})^{\top}] \\
    &= \beta^2 \sum_{i=0}^{n-1} (1-\beta)^{2i} \left( \E[\tilde\Delta] (\E[\tilde\Delta])^{\top} + \Cov[\tilde\Delta] \right) \\
    &\quad + 2\beta^2 \sum_{i,j=0: i\neq j}^{n-1} (1-\beta)^i (1-\beta)^j \E[\tilde\Delta] (\E[\tilde\Delta])^{\top}, \\
    &= \E[\mvae^{(t+n)}]  (\E[\mvae^{(t+n)}])^{\top} + \beta^2 \sum_{i=0}^{n-1} (1-\beta)^{2i} \Cov[\tilde\Delta].
\end{align*}
\normalsize
Therefore,
\small
\begin{align*}
    \Cov[\mvae^{(t+n)}] &= \E[\mvae^{(t+n)} (\mvae^{(t+n)})^{\top}] - \E[[\mvae^{(t+n)}] ([\mvae^{(t+n)}])^{\top} \\
    &= \beta^2 \sum_{i=0}^{n-1} (1-\beta)^{2i} \Cov[\tilde\Delta].
\end{align*}
\normalsize
Here,
\small
\begin{align*}
    \sum_{i=0}^{n-1} (1-\beta)^{2i} = \frac{1-(1-\beta)^{2n}}{1-(1-\beta)^2} = \frac{1-(1-\beta)^{2n}}{\beta(2-\beta)}.
\end{align*}
\normalsize
Thus, by ignoring $(1-\beta)^{2n}$, we can approximate $\Cov[\mvae^{(t+n)}]$ as follows:
\small
\begin{align*}
    \Cov[\mvae^{(t+n)}] &= [1-(1-\beta)^{2n}] \frac{\beta}{2-\beta} \Cov[\tilde\Delta], \\
    &\approx \frac{\beta}{2-\beta} \Cov[\tilde\Delta].
\end{align*}
\normalsize
Therefore, $\mvae^{(t+n)}$ approximately follows the distribution
\small
\begin{align*}
    \mvae^{(t+n)} 
    \sim \mathcal{D}\left( \E[\tilde\Delta], \frac{\beta}{2 - \beta} \Cov[\tilde\Delta]\right) .
\end{align*}
\normalsize
This completes the derivation of Eq.~(12).

\subsection{Derivation of Estimates for $\| \E[\tilde\Delta] \|_2^2$}
We organize the relation between $\mvae$ and $\tilde\Delta$ by the following equation:
\begin{align*}
\small
    \E[\| \mvae \|_2^2] &= \E[\mvae]^{\top} I \E[\mvae] + \Tr(\Cov[\mvae]) \nonumber \\
    &\approx \| \E[\tilde\Delta] \|_2^2 + \Tr \left( \frac{\beta}{2-\beta} \Cov[\tilde\Delta] \right) \nonumber  \\
    &= \| \E[\tilde\Delta] \|_2^2 + \frac{\beta}{2-\beta} \Tr(\Cov[\tilde\Delta]).
\end{align*}
\normalsize

Now we apply the same arguments to $\mvav$ and obtain
\begin{align*}
\small
    \E[\mvav] &= [1-(1-\beta)^{t+1}] \E[ \| \tilde\Delta \|_2^2 ] \nonumber \\
    &\approx \E[\| \tilde\Delta \|_2^2] = \| \E[\tilde\Delta] \|_2^2 + \Tr(\Cov[\tilde\Delta]).
\end{align*}
\normalsize
By reorganizing these arguments, we obtain
\begin{align*}
\small
    \| \E[\tilde\Delta] \|_2^2 \approx \frac{2-\beta}{2-2\beta} \E[\| \mvae \|_2^2] - \frac{\beta}{2-2\beta} \E[ \mvav ].
\end{align*}
\normalsize
This gives the rationale of the estimates $\frac{2-\beta}{2 - 2\beta} \norm{\mvae}_2^2 - \frac{\beta}{2-2\beta} \mvav$ for $\| \E[\tilde\Delta] \|_2^2$.

\section{Additional Experiment Results}
\label{sec:app_expresults}

\begin{figure*}[tb]
  \centering
  \includegraphics[width=170mm]{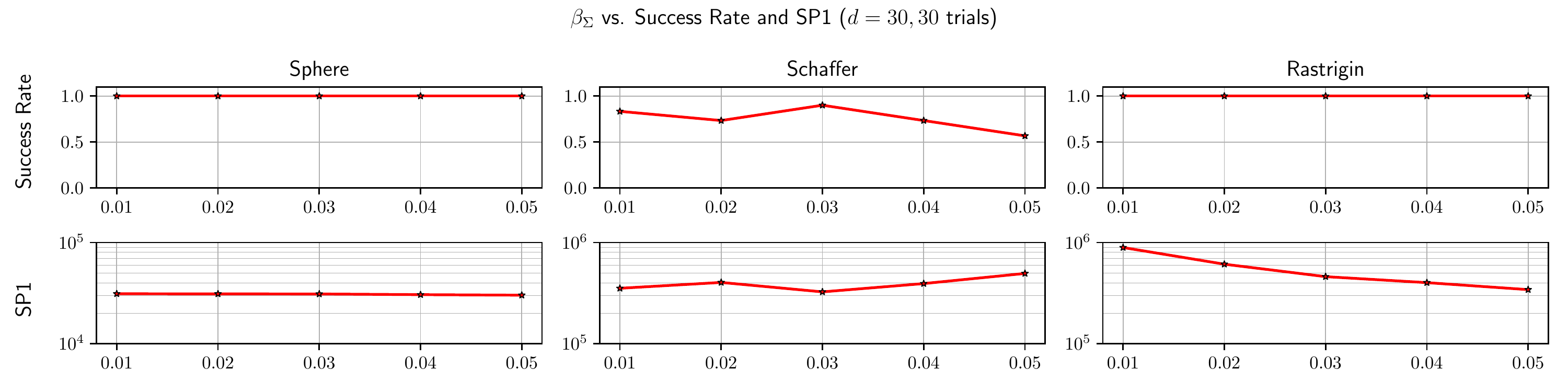}
  \caption{Success rate and SP1 versus hyperparameter $\beta_{\Sigma} \in \{ 0.01, 0.02, ..., 0.05 \}$  on 30-D noiseless problems.}
  \label{fig:beta_Sigma_srsp1_d=30_detailed}
\end{figure*}

Figure~\ref{fig:beta_Sigma_srsp1_d=30_detailed} shows the success rate and SP1 results with respect to $\beta_{\Sigma} \in \{ 0.01, 0.02, ..., 0.05 \}$ on the 30-D noiseless Sphere, Schaffer, and Rastrigin functions.
Clearly, the performance was not significantly affected by $\beta_{\Sigma}$ values within this range. However, similar to the case shown in Figure~\ref{fig:beta_Sigma_srsp1_d=30}, an excessively small $\beta_{\Sigma}$ setting decelerated the convergence for the Rastrigin function.

\begin{figure*}[tb]
  \centering
  \includegraphics[width=170mm]{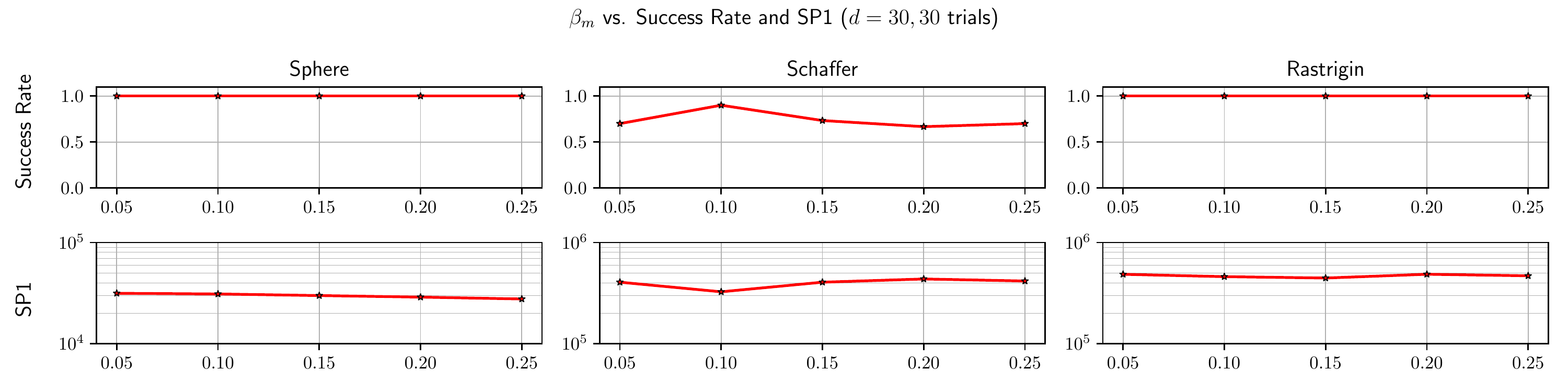}
  \caption{Success rate and SP1 versus hyperparameter $\beta_m$ on 30-D noiseless problems.}
  \label{fig:beta_mean_srsp1_d=30}
\end{figure*}

\begin{figure*}[htb]
  \centering
  \includegraphics[width=170mm]{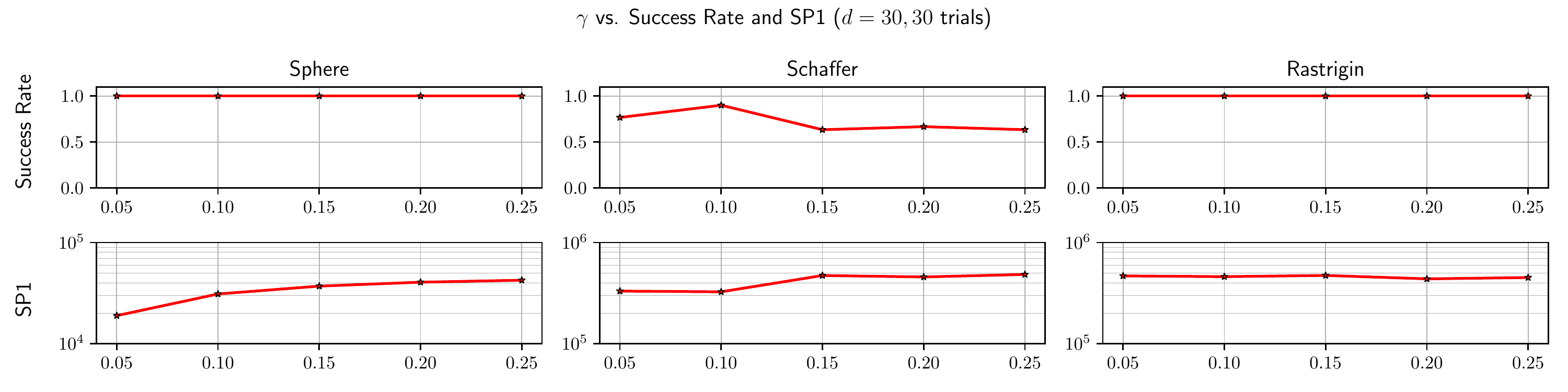}
  \caption{Success rate and SP1 versus hyperparameter $\gamma$ on 30-D noiseless problems.}
  \label{fig:gamma_srsp1_d=30}
\end{figure*}

Figures~\ref{fig:beta_mean_srsp1_d=30} and \ref{fig:gamma_srsp1_d=30} show the success rate and SP1 values with respect to $\beta_m$ and $\gamma$, respectively.
The results show that the performance was relatively stable against these hyperparameters.

\end{document}